\documentclass[10pt,twocolumn,letterpaper]{article}

\usepackage{iccv}
\usepackage{times}
\usepackage{epsfig}
\usepackage{graphicx}
\usepackage{amsmath}
\usepackage{amssymb}

\usepackage{multirow}
\usepackage{booktabs}
\usepackage{rotating}
\usepackage{color}
\usepackage[table]{xcolor}

\usepackage[pagebackref=true,breaklinks=true,letterpaper=true,colorlinks,bookmarks=false]{hyperref}

\iccvfinalcopy 

\def\iccvPaperID{5726} 

\ificcvfinal\fi

\begin{document}

\title{Fractal Pyramid Networks}

\author {
Zhiqiang Deng, \textsuperscript{\rm 1}
Huimin Yu, \textsuperscript{\rm 1,2}
Yangqi Long \textsuperscript{\rm 1} \\
\textsuperscript{\rm 1} Department of Information Science and Electronic Engineering, Zhejiang University, China \\
\textsuperscript{\rm 2} State Key Laboratory of CAD \& CG, China \\
{\tt\small \{zhichiang,yhm2005,longyangqi\}@zju.edu.cn}
}

\maketitle
\ificcvfinal\thispagestyle{empty}\fi

\begin{abstract}
  We propose a new network architecture, the Fractal Pyramid Networks (PFNs) for pixel-wise prediction tasks as an alternative to the widely used encoder-decoder structure. In the encoder-decoder structure, the input is processed by an encoding-decoding pipeline that tries to get a semantic large-channel feature. Different from that, our proposed PFNs hold multiple information processing pathways and encode the information to multiple separate small-channel features. On the task of self-supervised monocular depth estimation, even without ImageNet pretrained, our models can compete or outperform the state-of-the-art methods on the KITTI dataset with much fewer parameters. Moreover, the visual quality of the prediction is significantly improved. The experiment of semantic segmentation provides evidence that the PFNs can be applied to other pixel-wise prediction tasks, and demonstrates that our models can catch more global structure information. Our source codes are available in the supplementary materials.

\end{abstract}

\section{Introduction}

The encoder-decoder structure has been widely adopted in many pixel-wise prediction tasks~\cite{mayer2016large,godard2019digging,chen2018encoder,lin2017feature}. The encoder is utilized to encoding the information to a feature that has a large channel and the smallest resolution, and the decoder restores the spatial resolution from the high-level coded features. In this structure, the information is processed in an encoding-decoding pipeline. Some works~\cite{ronneberger2015u,fang2020towards} use skip-connection to transmit detailed information to the decoder and help the decoder to generate higher resolution output, however, the skip-connection cannot be considered as an encoding path since it does not generate a higher-level semantic feature.

Nowadays the encoder-decoder structure is facing many challenges and some of these challenges are caused by using the classification model as the backbone to get semantic features. The classification model is not properly designed for pixel-wise prediction tasks: (1) The classification model limits its ability to catch global context information, which is important to the pixel-wise prediction model~\cite{zhao2017pyramid,chang2018pyramid}. (2) Skip-connection is widely used in pixel-wise prediction models to help to recover the output resolution. But in this encoder-decoder structure, features in low-level are utilized to serve both the high-level feature generation and the resolution recovery, which may lead to sub-optimal results. Recently Fang~\etal~\cite{fang2020towards} also show that the improvement of the classification model may not boost the performance of the monocular depth estimation networks.

\begin{figure}
    \begin{center}
        \scalebox{0.3}[0.3]{\includegraphics{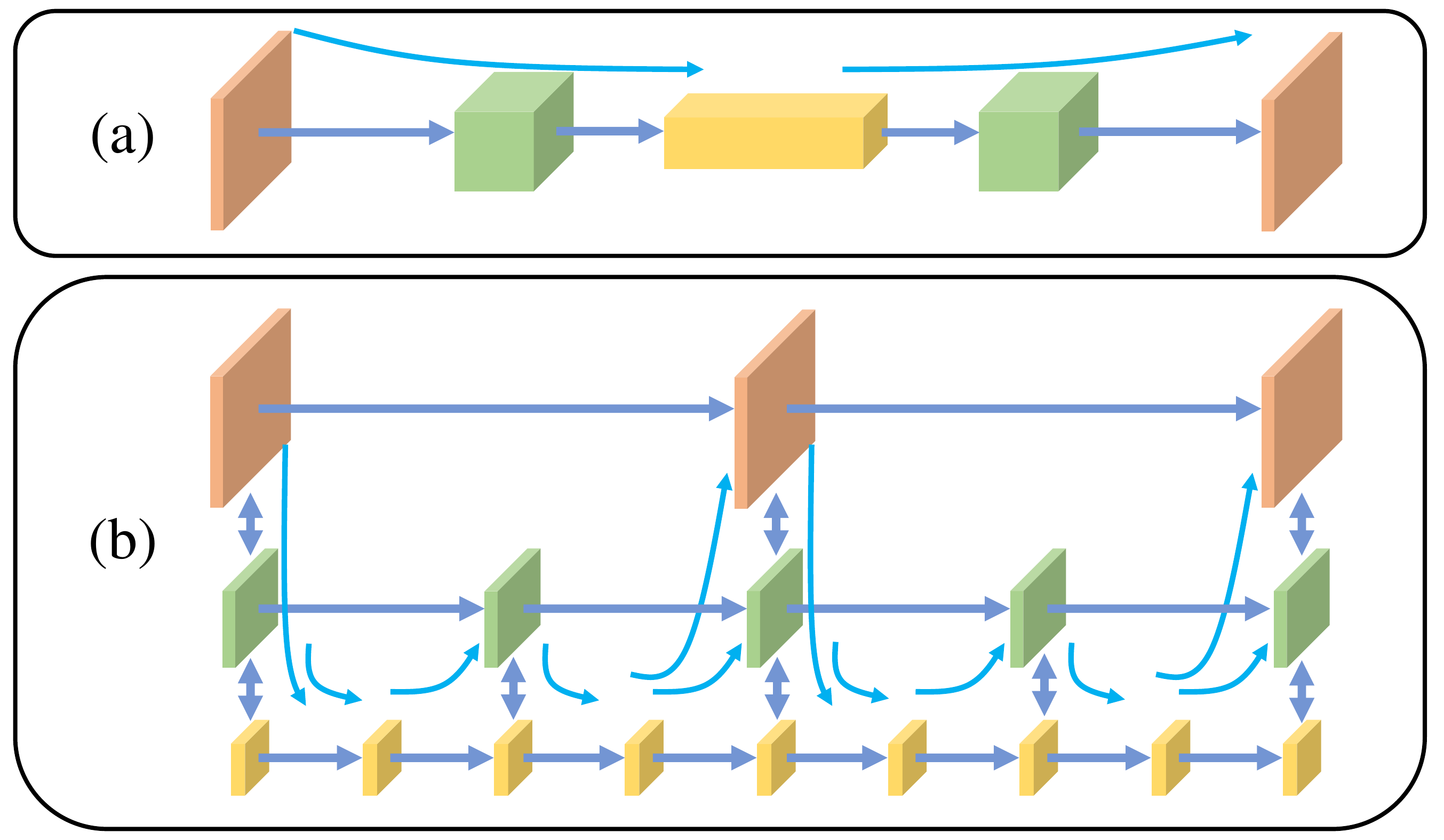}}
    \end{center}
    \caption{(a) A typical encoder-decoder structure. The information is processed in an encoding-decoding pipeline. (b) An example of our proposed PFNs. The PFN holds multiple information processing pathways. The colored cuboid represents the generated feature. The width of the cuboid represents the number of channels of the feature and the height represents the feature's resolution. The {\color[RGB]{0,176,240} blue} arrow represents the information processing pathway.}
    \label{Fig:Caption}
\end{figure}

In this paper, we fuse the pyramid structure into a fractal to form the fractal pyramid networks (PFNs) as an alternative to the encoder-decoder structure. By adopting the pyramid structure, the receptive field of the neuron in the small scale is enlarged to capture more global context information. Combining the pyramid and the fractal provides a new strategy to encode the information that rather than tries to get a large-channel semantic feature, we encode the information to multiple separate small-channel features by many signal processing pathways, as shown in Figure~\ref{Fig:Caption}. The features at different levels in the fractal are set to different resolutions to form the feature pyramids. Considering every scale of feature in the pyramid has its specific distribution, we separate the features in each scale into the private feature and shared feature, where the shared feature is used to communicate with other scales and the private feature is used to keep information in its own scale. We use the separation-and-aggregation module to automatically split and fuse the private feature and the shared feature. In this fractal, every feature in PFNs can be reached by another upstream feature in $O(scales)$ convolution layers without residual connection, and the middle features in PFNs are fully utilized for multiple signal processing pathways. Compared with mainstream pixel-wise prediction models, our models do not apply a modern classification model as the backbone and actually, our models do not have a backbone.

The proposed PFNs have the same form of input and output as the encoder-decoder structure, and so can be adopted to many pixel-wise prediction tasks like depth estimation, semantic segmentation, and surface normal prediction, etc. On the task of self-supervised monocular depth estimation, our model (1) can compete or outperform the state-of-the-art methods on the KITTI dataset with much fewer parameters, even without ImageNet~\cite{russakovsky2015imagenet} pretrained; (2) is much better than the un-pretrained ResNet18 backbone DepthNet, from $\mathrm{abs REL}$ 0.132 to 0.111; (3) predicts depth maps with higher visual quality, in which the shape is well reserved; (4) shows better temporal consistency when processing images from a video sequence. We also test our model on the semantic segmentation task in which the network is trained on virtual GTA5 dataset and straightforwardly evaluated on the Cityscapes dataset. The results (1) present the evidence that our models can be applied to other pixel-wise prediction tasks; (2) show that our models have better domain generalization ability which indicates that the PFN can catch more domain invariant global structure information; (3) demonstrate our proposed PFN restores the most details while maintaining satisfying global structure.

\section{Related Works}

\subsection{Image Classification}

The encoder-decoder structure is highly related to the image classification task which is a basic problem in the area of computer vision. Lots of efforts have been paid to improve the accuracy. AlexNet~\cite{krizhevsky2012imagenet} and VGGNets~\cite{simonyan2014very} stack the convolution layer to get the prediction. With the network going deeper, multi-branch convolutional networks were proposed to eases the difficulty of training networks with hundreds of layers. Inception models~\cite{szegedy2015going} demonstrates the benefits of increasing depth by carefully configure each branch with customized kernel filters. ResNets~\cite{he2016deep} designs a deeper network through the use of identity-based skip connections. SENets~\cite{hu2018squeeze} and SKNets~\cite{li2019selective} apply attention mechanism to further improve the performance. Among those networks, they all follow the 5-stage design that the feature downsamples the resolution to its half by pooling operation or stride convolution five times. In multi-branch convolutional networks, although there're multi-ways in the process of information encoding, they are following the same 5-stage design to encode the input information to a large channel feature which is of the smallest resolution. Recently, ViT~\cite{dosovitskiy2020image} applies a natural language processing model Transformer~\cite{vaswani2017attention} for the classification task. Despite its huge computation costs, how to apply it for pixel-wise prediction tasks is still underexplored.

\subsection{Encoder-Decoder Structure}

The encoder-decoder structure is widely used in computer vision. We focus its application on pixel-wise prediction tasks of depth estimation and semantic segmentation. In the area of depth estimation, DispNet~\cite{mayer2016large} stacks the convolution layer to encode semantic features and restore the output resolution by transpose convolution. To get more detailed information to recover the output resolution, more recent works~\cite{garg2016unsupervised,kuznietsov2017semi,godard2019digging,fang2020towards} adopted the skip-connection strategy to fuse low spatial resolution high-level feature map with high spatial resolution low-level feature map. In the area of semantic segmentation, FCNs~\cite{long2015fully} is proposed by adapting classification networks into fully convolutional networks. DeepLab series~\cite{chen2017deeplab,chen2017rethinking,chen2018encoder} uses atrous spatial pyramid pooling (ASPP). PSPNet~\cite{zhao2017pyramid} proposes to leverage the pyramid pooling module (PPM) to model multi-scale contexts. Among those encoder-decoder structure networks, they all follow the same design philosophy that high-level semantics are encoded in large-channel low-resolution feature maps.

\subsection{Fractal Networks}

While fractal architecture can be easily observed in nature, it attracts less attention in deep convolutional neural network design. FractalNet~\cite{larsson2016fractalnet} designs a fractal architecture for image classification. Our networks are related to this fractal architecture but have two important differences: (1) We introduce a new information encoding strategy that we encode the information to multiple small channel features, while in FractalNet the fractal architecture is used to form a processing block in the 5-stage design, which is still encoding the input information to a large channel feature. (2) We fuse the pyramid structure to the fractal architecture to form the whole network.

\begin{figure*}
    \begin{center}
        \scalebox{0.85}[0.85]{\includegraphics{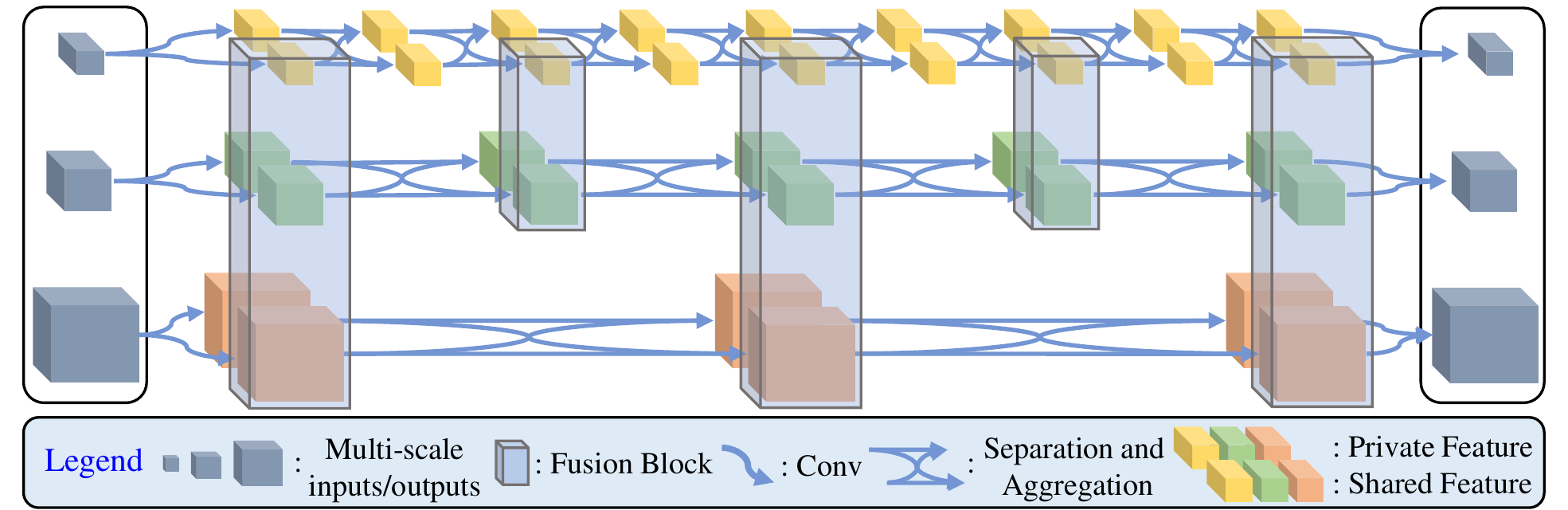}}
    \end{center}
    \caption{An example of our proposed fractal pyramid networks where the number of scales $S$ is set to 3. The colored cubes represent the features that the network generated. The PFN is composed of a fractal architecture in which the smallest scale has the deepest layers. We use the fusion block to fuse the shared features from each scale which acts as the bridge to transmit information between each row in this figure. The features in the fusion blocks are the shared features and the private features are behind the shared features.}
    \label{Fig:Archi}
\end{figure*}

\subsection{Pyramid Structure}

Many works adopt the pyramid structure to model multi-scale contexts. The aforementioned DeepLab series uses ASPP, and PSPNet proposes PPM for semantic segmentation. FPN~\cite{lin2017feature} exploit the inherent pyramidal hierarchy to construct feature pyramids and later be applied by Monodepth2~\cite{godard2019digging} for self-supervised monocular depth estimation. In stereo matching, PSMNet~\cite{chang2018pyramid} learns the relationship between an object and its sub-regions by the proposed spatial pyramid pooling module for stereo matching. In optical flow, SPyNet~\cite{ranjan2017optical} introduces image pyramids to estimate optical flow in a coarse-to-fine approach. PWC-Net~\cite{sun2018pwc} and RAFT~\cite{teed2020raft} improves optical flow estimation by using feature pyramids. Among those networks, they all apply pyramid operation in a few places, while our PFNs benefit from pyramid structure everywhere.

\section{Fractal Pyramid Networks}

\subsection{Network Overview}
In this section, we start to formally describe the fractal pyramid networks. Firstly, we define $Z_{S}=\{z_1, z_2, ..., z_{S}\}$ as a set of inputs with $S$ scales, $N_{S}=\{n_1, n_2, ..., n_{S}\}$ as a set of hyper-parameters, and define $s$ as the scale index of the fractal $f_{s}(\cdot)$. The basic case of $f_{s}(\cdot)$ is the $f_{1}(\cdot)$, defined as:
\begin{equation}
f_{1}(Z_1) = \{SA(z_1)\}
\end{equation}
where $SA(\cdot)$ denotes separation and aggregation module. Then we can define the successive fractals recursively:
\begin{equation}
\begin{aligned}
f_{s+1}(Z_{s+1}) &= FU(f_{s}^{n_s}(Z_s)\cup\{SA(z_{s+1})\}) \\
f_{s}^{n_s}(Z_{s}) &= \underbrace{f_{s}(f_{s}(...f_{s}(}_{n_{s}}Z_{s})...))
\end{aligned}
\end{equation}
where $FU(\cdot)$ denotes fusion block, which will be described in Sec.~\ref{Sec:FusBlock}. Note that there can be $n_{s}$ $f_{s}(\cdot)$s to compose the $f_{s+1}(\cdot)$, we set the $n_{s}$ a constant number $2$ for simplicity in this paper unless otherwise noted. Finally we can define our fractal pyramid network $p_{s}(\cdot)$ as:
\begin{equation}
p_{s}(Z_{s}) = f_{s}^{n_s}(Z_{s})
\end{equation}

\begin{figure}
\begin{center}
\scalebox{0.4}[0.4]{\includegraphics{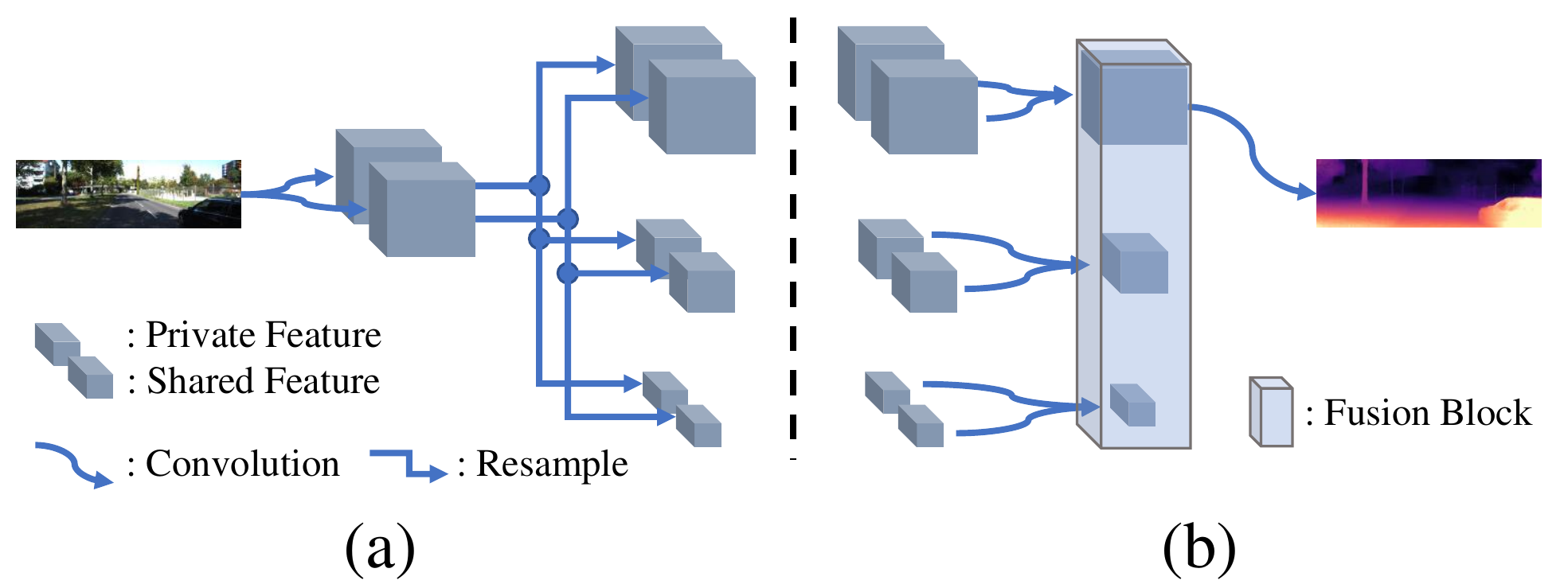}}
\end{center}
   \caption{Input and output design for PFNs. (a) We first use two convolution layer for the input image to generate the private and shared features, and then resample them to form the feature pyramid. (b) The shared and the private features are firstly concatenated and then be fused. Finally the fused feature are fed to a convolution layer to get the final prediction.}
\label{Fig:InOut}
\end{figure}
Figure~\ref{Fig:Archi} shows the case when $S=3$, as there are 3 rows in this figure. The PFN receives 3 scale inputs and predicts the corresponding 3 scale outputs. Note in this architecture, when we connect the inputs and the outputs, there are multiple information processing pathways. One can see that our PFN rather encodes signals in a large-channel feature, but encode them in separate small-channel features.

Many pixel-wise prediction tasks receive a single-scale image and just need single-scale output. In this case, as shown in Figure~\ref{Fig:InOut}, we firstly apply two convolution layers to the input to get the initial private feature and shared feature, and then form the pyramid features by recursively average pooling. For output, we concatenate the last shared feature and private feature of each scale to form a new shared feature, then input the new shared feature pyramid to the output fusion block, and finally apply a convolution layer to the wanted scale to get the prediction.

\subsection{Fusion Block}
\label{Sec:FusBlock}
\begin{figure}
\begin{center}
\scalebox{0.4}[0.4]{\includegraphics{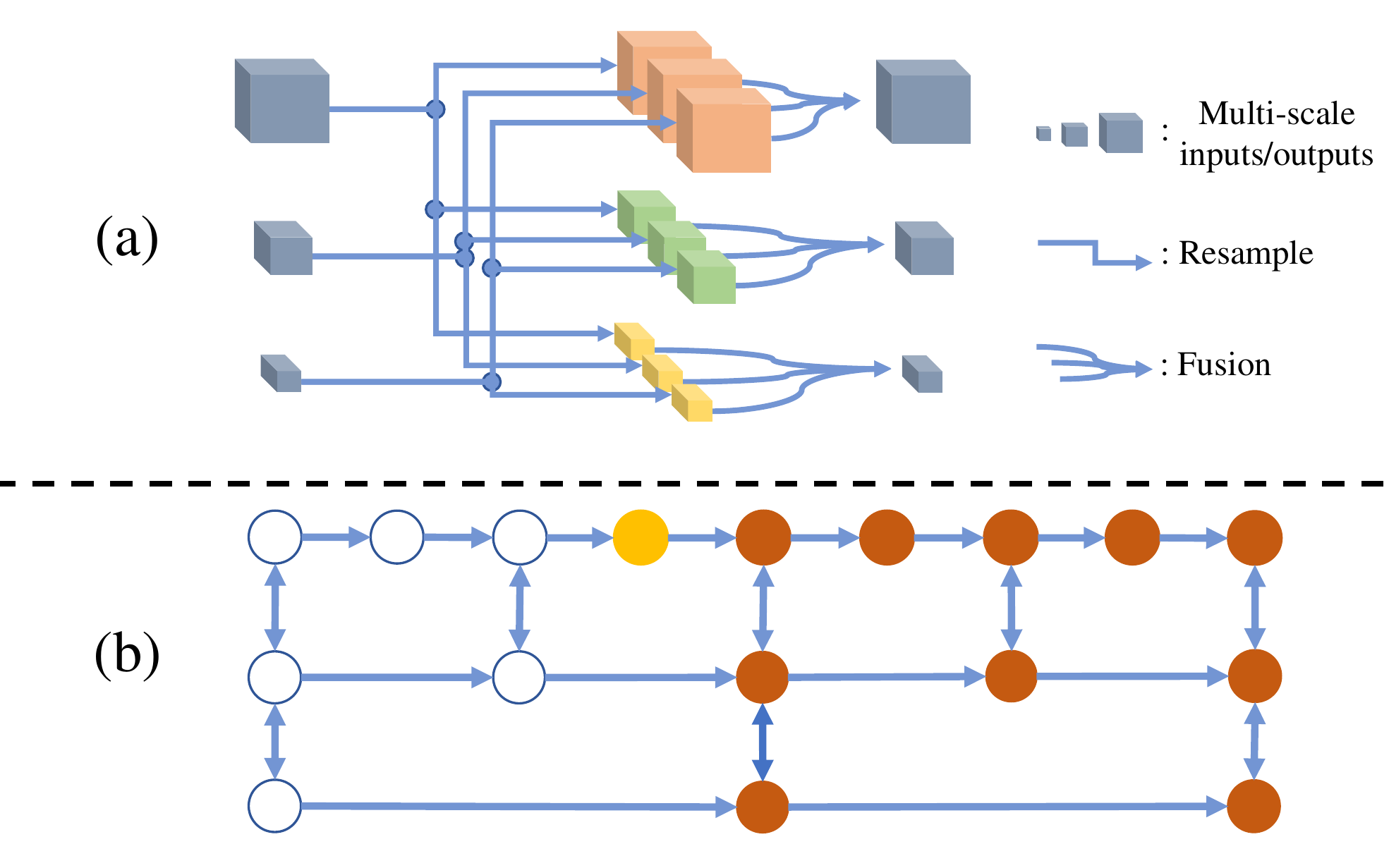}}
\end{center}
   \caption{(a) Fusion block. The fusion block takes multi-scale features as input and also output multi-scale features. Those input features are firstly resampled to each scale and then fused to generate the output features. (b) The middle feature (the {\color[RGB]{255,192,0}yellow} circle) can influence many downstream features (the {\color[RGB]{197,90,17}orange} circle).}
\label{Fig:FUPPG}
\end{figure}

To exchange signals between features in different scales, we design the fusion block to aggregate information from each scale. The fusion block performs as the "bridge" between scales to form multiple information processing pathways otherwise the PFNs collapse to multi-branch encoder-decoder networks. As shown in Figure~\ref{Fig:FUPPG} (a), to handle pyramid features, we firstly resample features into each specific scale, and then fuse those features per scale:
\begin{equation}
FU(Z_{s}) = \{\mathrm{fuse}(\mathrm{rescale}_a(Z_s))|a\in\{1,2,...,S\}\}
\end{equation}
where $\mathrm{fuse}(\cdot)$ denotes the fusion operation, and $\mathrm{rescale}_a(\cdot)$ denotes the sample operation to the target scale $a$. We propose two types of fusion operation: the channel-wise weighted sum (CWS) $\mathrm{fuse}_{\mathrm{cws}}$ and the concatenation to convolution (CTC) $\mathrm{fuse}_{\mathrm{ctc}}$. For CWS fusion, rather to simply add those features, we apply different weight to different channel for each scale feature:
\begin{equation}
\mathrm{fuse}_{\mathrm{cws}}(Z_s) = \{\mathrm{cat}(\sum_{i=1}^{L}{w_{si}*l_{si}})|s\in\{1,2,...,S\}\}
\end{equation}
where $L$ is the count of channel of $z_s$, $l_{si}$ is the channel $i$ of $z_s$, and $w_{si}$ is the weight for the $l_{si}$. $\mathrm{cat}(\cdot)$ denotes concatenation operation. For CTC fusion, we concatenate the features and then apply a convolution layer to fuse them:
\begin{equation}
\mathrm{fuse}_{\mathrm{ctc}}(Z_s) = \{\mathrm{conv}(\mathrm{cat}(z_s))|s\in\{1,2,...,S\}\}
\end{equation}
where $\mathrm{conv}(\cdot)$ denotes the convolution operation.

We test a series of combinations of fusion choices between the output fusion block and the other fusion blocks. We find that using CTC fusion in the output fusion block and using CWS fusion in other fusion blocks results in the best performance. So we choose the combination of (CWS, CTC) for experiments.

\subsection{Separation and Aggregation Module}
\label{Sec:SAModule}

The separation and aggregation (SA) module has been widely used in many networks, including cross-stitch networks~\cite{misra2016cross} and attentional separation-and-aggregation network~\cite{gao2020attentional}. In PFNs, considering that each scale has its specific distribution that not every feature in this scale is suitable for share to other scales, we separate the feature into a private one and a shared one. The shared features are used to exchange messages between different scales in fusion block, and the private features are used to keep their own peculiarity of this scale,~\eg features for resolution recovery. We use the SA module to handle them, in which the network can learn which feature to share and which feature to keep. Like the SA module that applied in other works, we form our SA module with several convolution layers:
\begin{equation}
\begin{aligned}
SA(z_s) &= \{\mathrm{conv}(\mathrm{cat}(z_s)), \mathrm{conv}(\mathrm{cat}(z_s))\} \\
z_s &= \{zs_{s}, zp_{s}\}
\end{aligned}
\end{equation}
where $zs_{s}$ denotes the shared feature and $zp_{s}$ denotes the private feature. Note that the private feature can have a different number of channels from the shared feature, we assign $sc$ as the number of channels for the shared feature, and $pc$ for the private feature.

\subsection{Additional Considerations}

The proposed PFNs hold multiple information processing pathways. As shown in Figure~\ref{Fig:FUPPG} (b), the middle feature in the network can be utilized by many downstream features. In other words, the middle feature has a big influence on those downstream features. In the extreme case, once the middle feature has an infinity output, those downstream features will output \verb'nan' as the infinity float cannot be calculated. A real neuron in the human brain also has its energy limits while the ReLU activation function does not handle this problem. In the PFNs, we simply clip the outputs of the SA module to range $[0, 10^4]$ to avoid the "nan" phenomenon and enhance the robustness.

Although every feature in PFNs can be reached by any upstream feature in $O(scale)$ convolution layers, however, when we set $S$ too large,~\eg $S=7$, lots of features in the smallest scale still need to get through many convolution layers to reach the output. The long convolution path makes the gradient explosion easily occurs which decreases the performance~\cite{bengio1994learning}. We clip the gradient norm to range $[0, 1]$ in the backward process using the \verb'clip_grad_norm_' tool to stabilizing the training process.

\section{Experiments on Self-supervised Monocular Depth Estimation}
Monocular Depth Estimation (MDE) is an essential pixel-wise prediction problem that serves various real-world applications like autonomous vehicles, robotic navigation, and augmented reality, etc. However, learning monocular depth via direct supervision requires accurate and large ground-truth datasets from additional sensors and precise cross-calibration which is a major challenge for the further development of the area. Self-supervised methods can overcome those limitations that they just need video sequences or multi-view images as their supervision. Compare with the ground truth depth, these data are easier to collect. In this experiment, we follow the settings of Monodepth2~\cite{godard2019digging} that use three adjacent frames in a video sequence to train our networks.

\subsection{Problem Formulation}
Given an Image $I_t$, our PFN model $f_D$ is applied to predict the scale-ambiguous depth $\hat{d}=f_D(I_t(p))$ for every pixel $p$ in the target image $I_t$. In this self-supervised MDE, we need the PoseNet $f_P$ to estimate the camera relative pose $[R,t]$ of the camera from the source image $I_s$ to the target image $I_t$. As we get the estimated depth map $\hat{d}$ and the pose $[R,t]$, a geometry constraint can be applied to form the appearance matching loss $\mathcal{L}_{\mathrm{ap}}(I_t, \hat{I}_{s\to t})$:
\begin{equation}
\mathcal{L}_{\mathrm{ap}}(I_t, \hat{I}_{s\to t}) = \mathcal{L}_{\mathrm{p}}(I_t, \hat{I}_{s\to t})\odot\mathcal{M}_t\odot\mathcal{M}_p
\end{equation}
Here, $\hat{I}_{s\to t}$ is the warped source image according to the predicted depth $\hat{d}$ to predict the target image $I_t$. Photometric loss $\mathcal{L}_{\mathrm{p}}$ calculate the pixel-level similarity between the target image $I_t$ and the synthesized image $\hat{I}_{s\to t}$ using the Structural Similarity (SSIM) and L1 distance:
\begin{equation}
\mathcal{L}_{\mathrm{p}}(a, b) = \alpha\cdot\left\|a-b\right\|_1 +
 (1 - \alpha)\cdot\frac{1-\mathrm{SSIM}(a, b)}{2}
\end{equation}
where $\alpha=0.85$. As in Monodepth2~\cite{godard2019digging}, we calculate the $\mathcal{M}_t$ to mask out the pixels that do not have a valid mapping by select the minimum loss around the photometric losses for training:
\begin{equation}
\mathcal{L}_{\mathrm{p}} = \min_s\mathcal{L}_{\mathrm{p}}(I_t, \hat{I}_{s\to t})
\end{equation}
and $\mathcal{M}_p$ is utilized to exclude static pixels pixels that have a warped photometric loss $\mathcal{L}_{\mathrm{p}}(I_t, \hat{I}_{s\to t})$ higher than their corresponding unwarped photometric loss $\mathcal{L}_{\mathrm{p}}(I_t, I_s)$:
\begin{equation}
\mathcal{M}_p = \min_s\mathcal{L}_{\mathrm{p}}(I_t, \hat{I}_{s\to t}) > \min_s\mathcal{L}_{\mathrm{p}}(I_t, I_s)
\end{equation}

Since many probabilities satisfy the constraint, we utilize the contrast-sensitive smooth loss to train the networks:
\begin{equation}
\mathcal{L}_{\mathrm{s}}(d, I) = \left|\partial_x d^{*}\right|e^{-\left|\partial_x I\right|} +
\left|\partial_y d^{*}\right|e^{-\left|\partial_y I\right|}
\end{equation}
where $d^{*}=d/\bar{d}$ is the mean-normalized inverse depth.
The total loss for training is:
\begin{equation}
\mathcal{L} = \mathcal{L}_{\mathrm{ap}} + \gamma\mathcal{L}_{\mathrm{s}}
\end{equation}

\subsection{Experiment Setup}

\noindent\textbf{KITTI Dataset}~\cite{geiger2013vision}. We train and evaluate our models on the KITTI 2015 dataset which contains videos in 200 street scenes captured by RGB cameras, with sparse depth ground truths captured by Velodyne laser scanner. For fair comparisons, we adopt the training protocol used in Eigen~\etal~\cite{eigen2015predicting} and follow Zhou~\etal's~\cite{zhou2017unsupervised} pre-processing to remove static frames. This results in 39810 monocular triplets for training, 4424 for validation, and 697 for evaluation. Due to GPU memory limits, we resize the input images to $640\times192$. When testing, we upsample the output depth to the original resolution for evaluation.

\noindent\textbf{Implementation Details}. We straightforwardly apply our PFNs as the DepthNet to predict depth from a single image. We set the scale $S$ to 5 as the receptive field of the smallest feature is big enough for the convolution layer to grab global information. The $sc$ and $pc$ are set to $18$ and $54$ respectively. We follow the Monodepth2~\cite{godard2019digging} that applies losses to 4 scales for the outputs. As our models are different from encoder-decoder networks, we do not use the ImageNet pretrained image classification model as the backbone to predict semantic features.

\begin{table*}[htbp]
  \centering
  \setlength{\tabcolsep}{4.0pt}
  \caption{Quantitative performance comparison of the PFN on the KITTI dataset for distances up to 80m. $\delta_i$ denotes metric $\delta<1.25^i$. The best results are in \textbf{bold}, and the second-best are \underline{underlined}. Pretrained means the backbone is pretrained on the ImageNet dataset.}
    \begin{tabular}{c|l|c|r|cccc|ccc}
    \hline
    \multirow{2}[2]{*}{} & \multicolumn{1}{c|}{\multirow{2}[2]{*}{Method}} & \multicolumn{1}{c|}{\multirow{2}[2]{*}{Resolution}} & \multicolumn{1}{c|}{\multirow{2}[2]{*}{Params}} & \multicolumn{4}{c|}{\cellcolor[rgb]{.988,.894,.839}The lower the better} & \multicolumn{3}{c}{\cellcolor[rgb]{.886,.937,.855}The higher the better} \\
          &       &       &       & \cellcolor[rgb]{.988,.894,.839}Abs REL & \cellcolor[rgb]{.988,.894,.839}Sq REL & \cellcolor[rgb]{.988,.894,.839}RMSE  & \cellcolor[rgb]{.988,.894,.839}$\text{RMSE}_{\log}$ & \cellcolor[rgb]{.886,.937,.855}$\delta_1$  & \cellcolor[rgb]{.886,.937,.855}$\delta_2$ & \cellcolor[rgb]{.886,.937,.855}$\delta_3$ \\
    \hline
    \hline
    \multirow{6}[4]{*}{\begin{sideways}w. pretrained\end{sideways}} & DualNet~\cite{zhou2019unsupervised} & 1248x384 & \multicolumn{1}{l|}{25.56M} & 0.121 & 0.837 & 4.945 & 0.197 & 0.853 & 0.955 & 0.982 \\
          & DeFeat-Net~\cite{spencer2020defeat} & 480x352 & \multicolumn{1}{l|}{14.84M} & 0.126 & 0.925 & 5.035 & 0.200 & 0.862 & 0.954 & 0.980 \\
          & Zhao~\etal~\cite{zhao2020towards} & 832x256 & \multicolumn{1}{l|}{14.84M} & 0.113 & \textbf{0.704} & \textbf{4.581} & \textbf{0.184} & 0.871 & \underline{0.961} & \textbf{0.984} \\
          & Johnston~\etal~\cite{johnston2020self} & 640x192 & \multicolumn{1}{l|}{44.55M} & \textbf{0.106} & 0.861 & 4.699 & \underline{0.185} & \textbf{0.889} & \textbf{0.962} & 0.982 \\
\cline{2-11}          & Monodepth2 (ResNet18)~\cite{godard2019digging} & 640x192 & \multicolumn{1}{l|}{14.84M} & 0.115 & 0.903 & 4.863 & 0.193 & 0.877 & 0.959 & 0.981 \\
          & Monodepth2 (ResNet50) & 640x192 & \multicolumn{1}{l|}{34.57M} & \underline{0.111}  &  \underline{0.789}  &  \underline{4.621}  &  0.186  &  \underline{0.879}  &  \textbf{0.962}  & \underline{0.983} \\
    \hline
    \hline
    \multirow{11}[6]{*}{\begin{sideways}w/o. pretrained\end{sideways}} & SfMLearner~\cite{zhou2017unsupervised} & 416x128 & \multicolumn{1}{l|}{31.60M} & 0.208 & 1.768 & 6.856 & 0.283 & 0.678 & 0.885 & 0.957 \\
          & Vid2Depth~\cite{mahjourian2018unsupervised} & 416x128 & \multicolumn{1}{l|}{31.60M} & 0.163 & 1.24  & 6.220 & 0.250 & 0.762 & 0.916 & 0.968 \\
          & DF-Net~\cite{zou2018df} & 576x160 & \multicolumn{1}{l|}{25.56M} & 0.150 & 1.124  & 5.507 & 0.223 & 0.806 & 0.933 & 0.973 \\
          & Struct2Depth~\cite{casser2019depth} & 416x128 & \multicolumn{1}{l|}{31.60M} & 0.141 & 1.026 & 5.291 & 0.215 & 0.816 & 0.945 & 0.979 \\
          & DualNet~\cite{zhou2019unsupervised} & 1248x384 & \multicolumn{1}{l|}{25.56M} & 0.135 & 0.973 & 5.235 & -     & 0.823 & 0.947 & 0.980 \\
          & Zhao~\etal~\cite{zhao2020towards} & 832x256 & \multicolumn{1}{l|}{14.84M} & 0.130 & 0.893 & 5.062 & 0.205 & 0.832 & 0.949 & 0.981 \\
          & PackNet-SfM~\cite{guizilini20203d} & 640x192 & \multicolumn{1}{l|}{128.3M} & \textbf{0.111} & \underline{0.785} & \underline{4.601} & \underline{0.189} & \textbf{0.878} & \underline{0.960} & \underline{0.982} \\
\cline{2-11}          & Monodepth2 (ResNet18)~\cite{godard2019digging} & 640x192 & \multicolumn{1}{l|}{14.84M} & 0.132 & 1.044 & 5.142 & 0.210 & 0.845 & 0.948 & 0.977 \\
          & Monodepth2 (ResNet50) & 640x192 & \multicolumn{1}{l|}{34.57M} &  \underline{0.126}  &  0.950  &  4.984  &  0.202  &  0.854  &  0.952  & 0.979 \\
          & Monodepth2 (DispNet) & 640x192 & \multicolumn{1}{l|}{31.60M} & 0.136  & 1.080  & 5.357  &  0.217 &  0.837 &  0.942 & 0.975 \\
\cline{2-11}          & \cellcolor[rgb]{.851,.851,.851} PFN (sc=18,pc=54) & \cellcolor[rgb]{.851,.851,.851} 640x192 &\multicolumn{1}{l|}{\cellcolor[rgb]{.851,.851,.851}4.824M} & \cellcolor[rgb]{.851,.851,.851} \textbf{0.111}  & \cellcolor[rgb]{.851,.851,.851} \textbf{0.761}  &  \cellcolor[rgb]{.851,.851,.851}\textbf{4.569} &  \cellcolor[rgb]{.851,.851,.851}\textbf{0.186}  & \cellcolor[rgb]{.851,.851,.851}\underline{0.877}   &  \cellcolor[rgb]{.851,.851,.851}\textbf{0.961} & \cellcolor[rgb]{.851,.851,.851}\textbf{0.983} \\
    \hline
    \end{tabular}%
  \label{Tab:Compare}%
\end{table*}%

For PoseNet, we adopt a modified ResNet-18~\cite{he2016deep} structure that accepts a concatenated image pair for input. As in~\cite{shu2020feature}, the PoseNet that uses a light-weight backbone has pose accuracy similar to the one that uses a backbone with more parameters. So we use the light-weight backbone to reduce computation complexity and save memory.

We use the PyTorch library with all models trained on an NVIDIA GeForce RTX 3090 GPU with 24GB memory. The networks are trained for 50 epochs with a batch size of 4. The learning rate is set to $1\times 10^{-4}$ and keeps stable during the training process. All of our models are trained with Adam optimizer with $\beta_1=0.9$ and $\beta_2=0.999$. We set the depth regularization weight to $\gamma = 0.001$.

\subsection{Comparison With the Encoder-Decoder Structure}

We compare our models with encoder-decoder structures which are proposed for self-supervised monocular depth estimation. We add sigmoid layers for the last convolution outputs of our models as the necessary modification without any other changes. Noticing that there are many frameworks~\cite{zhao2020towards,godard2019digging} are used, we select the framework used in Monodepth2~\cite{godard2019digging} as it has been adopted by many current arts~\cite{guizilini20203d,spencer2020defeat}. For better comparison, we choose ResNet50~\cite{he2016deep} and DispNet~\cite{mayer2016large} as the different backbones for the DepthNet used in Monodepth2. We also report the number of parameters of those networks. We only count the parameters of the DepthNet which is used for depth prediction. The PoseNet and any other networks are excluded for counting. As some works have not published their code, we simply use the backbone of their DepthNet for parameter counting.

\begin{figure*}
\begin{center}
\scalebox{0.3}[0.3]{\includegraphics{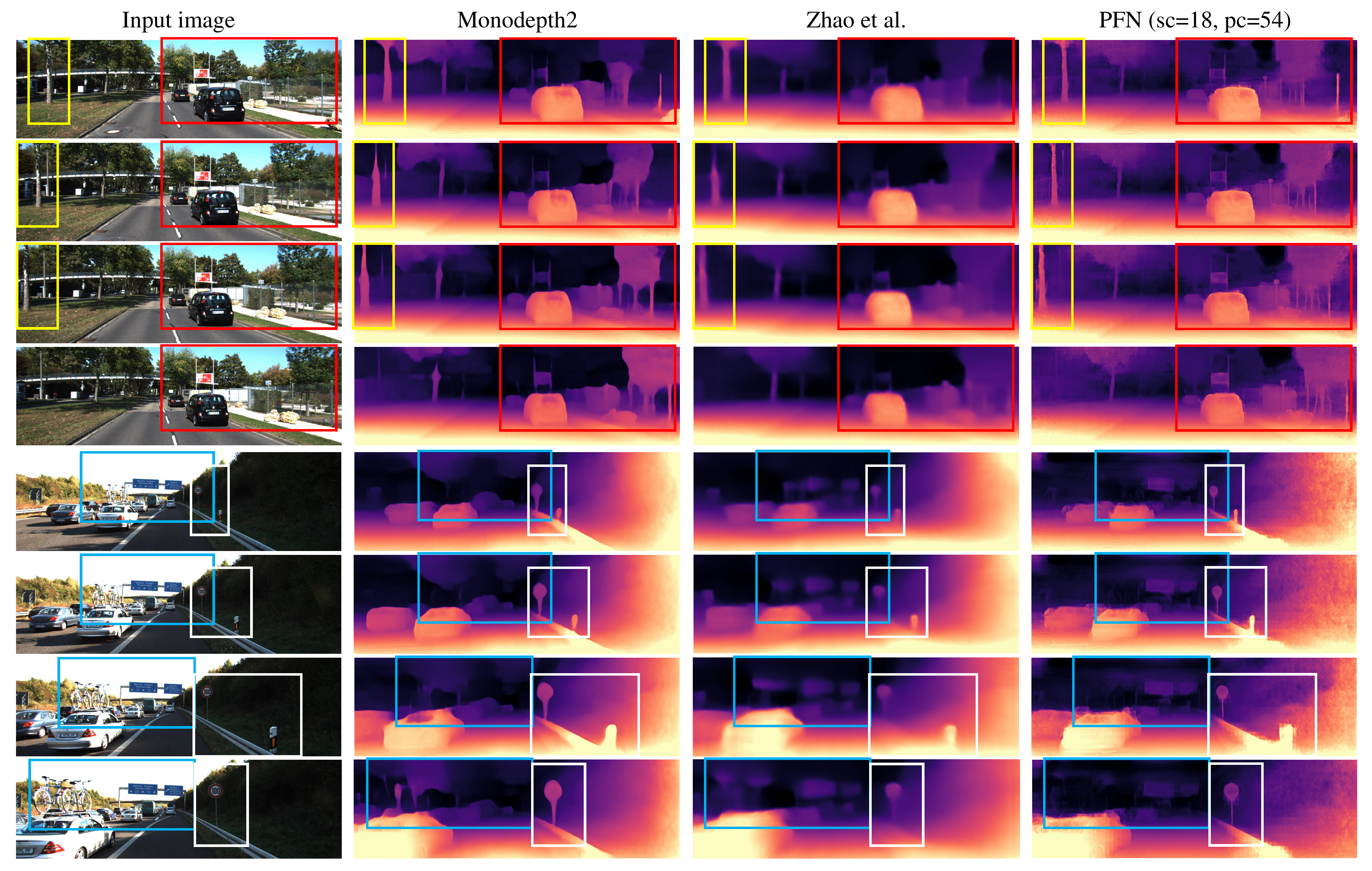}}
\end{center}
   \caption{Qualitative comparison between our PFN and previous methods on frames from the KITTI dataset (Eigen test split). Our network is able to capture more accurate shape and predicts images in a video sequence with good temporal consistency.}
\label{Fig:DepthViz}
\end{figure*}

Table~\ref{Tab:Compare} shows the results in which our models get comparable results with the state-of-the-art. Without ImageNet pretrained, the proposed PFN is much better than the Monodepth2 with ResNet18 backbone, from $\mathrm{abs REL}$ 0.132 to 0.111. Moreover, our model gets the best $\mathrm{RMSE}$ around those networks whether they are pretrained or not. Note that the parameters of our model are $3.1\times$ smaller than the DepthNet with ResNet18 and $6.6\times$ smaller than the DispNet, this indicates that our model has a stronger representation ability. Although the PackNet-SfM~\cite{guizilini20203d} gets a better $\delta_1$, our model outperforms it in the rest of metrics despite the parameters of the PackNet-SfM is $26.6\times$ more than our model's. Figure~\ref{Fig:DepthViz} qualitatively illustrates the performance of our model, in which our model gets significantly better visual quality. The shape of the object (\eg the thin poles and the traffic signs) is well reserved.

\subsection{Temporal Consistency}

In real-world applications like automatic video special effects and robots, a stable and consistent output from the network is desired. Many methods~\cite{zhang2019exploiting,luo2020consistent} have been proposed to handle this issue where some of them need multi-frame inputs. Lei~\etal~\cite{lei2020blind} hypothesize that the flickering artifacts in a video are caused by overfitting. To quantitatively evaluate the temporal consistency, we propose two metrics named Temporal Absolute Consistency (TAC) and Temporal Relative Consistency (TRC). They are defined as:
\begin{equation}
\begin{aligned}
    \text{TAC}&: \left|\hat{d}_{t}-\hat{d}_{t+1\to t}\right|\qquad
    \text{TRC}&: \frac{\left|\hat{d}_{t}-\hat{d}_{t+1\to t}\right|}{\max{\hat{d}_{t},\hat{d}_{t+1\to t}}}
\end{aligned}
\end{equation}
where $\hat{d}_{t}$ is the prediction of frame $t$ and $\hat{d}_{t+1\to t}$ is the warped prediction of frame $t+1$. We use ground truth flow provided by KITTI 2015 with 200 pairs of images to warp the prediction. Since sparse ground truth flows are provided, we only calculate those pixels with valid flow values. Table~\ref{Tab:TemCon} shows the results in which our model is significantly better. We conjecture that this is because our model holds fewer parameters and can recover more accurate shapes. Figure~\ref{Fig:DepthViz} visualize our model's advantages, in which our model gets predictions of better temporal consistency when processing images from a video sequence.

\begin{table}[htbp]
  \setlength{\tabcolsep}{13.2pt}
  \centering
  \footnotesize
  \caption{Quantitative analysis of temporal consistency. \dag indicates ImageNet pretraining. $\downarrow$ denotes lower is better. We compare our model with Monodepth2 and Zhao~\etal. We also report the results of Monodepth2 when using larger backbone for better comparison.}
  \vspace{2pt}
    \begin{tabular}{l|cc}
    \hline
    Method & \cellcolor[rgb]{.988,.894,.839}$\mathrm{TAC}\downarrow$   & \cellcolor[rgb]{.988,.894,.839}$\mathrm{TRC}\downarrow$ \\
    \hline
    \hline
    Monodepth2 (DispNet) & 2.2290 &  0.1334 \\
    Monodepth2 (ResNet 18)~\cite{godard2019digging} \dag &   0.8547    &  0.1316 \\
    Monodepth2 (ResNet 50) \dag &   1.0630    &  0.1344 \\
    Zhao~\etal~\cite{zhao2020towards} \dag &    0.7734   & 0.1339  \\
    \hline
    \rowcolor[rgb]{.851,.851,.851} PFN (sc=18, pc=54) &   \textbf{0.7373}    &  \textbf{0.1284} \\
    \hline
    \end{tabular}%
  \label{Tab:TemCon}%
\end{table}%

\begin{table}[htbp]
  \centering
  \footnotesize
  \setlength{\tabcolsep}{2.5pt}
  \caption{Ablation study on the SA module. $pc=0$ means we do not use SA module in the PFNs.}
  \vspace{2pt}
    \begin{tabular}{ccc|ccccc}
    \hline
    shared & private & sc:pc &\cellcolor[rgb]{.988,.894,.839} Abs REL &\cellcolor[rgb]{.988,.894,.839} Sq REL &\cellcolor[rgb]{.988,.894,.839} RMSE  & $\cellcolor[rgb]{.988,.894,.839} \text{RMSE}_{\log}$ &\cellcolor[rgb]{.886,.937,.855} $\delta_1$ \\
    \hline
    \hline
    48 & 0  & 1:0    & 0.116 & 0.839 & 4.660 & 0.190 & 0.871 \\
    32 & 32 & 1:1    & 0.112 & 0.776 & 4.574 & 0.188 & 0.875 \\
    24 & 48 & 1:2    & 0.113 & 0.778 & 4.564 & 0.187 & 0.873 \\
    18 & 54 & 1:3    & 0.111 & 0.761 & 4.569 & 0.186 & 0.877 \\
    14 & 56 & 1:4    & 0.116 & 0.754 & 4.666 & 0.192 & 0.864 \\
    \hline
    \end{tabular}%
  \label{Tab:EffSA}%
\end{table}%

\subsection{Ablation Study}

\noindent\textbf{Effect of separation and aggregation module.} We use the separation and aggregation module to decouple the domain-specific feature and the shared feature. To verify the effect of the SA module, we test a series of hyper-parameters that set different values to the private channel $pc$ and the shared channel $sc$ while keeping the number of parameters roughly the same. Note that when $pc$ is set to $0$, this means all the features from different scales are shared. As shown in Table~\ref{Tab:EffSA}, the SA module significantly improves the performance of the PFN. We also found the case $sc:pc=1:3$ brings the best results comparing with other settings.

\begin{figure}
\begin{center}
\scalebox{0.20}[0.20]{\includegraphics{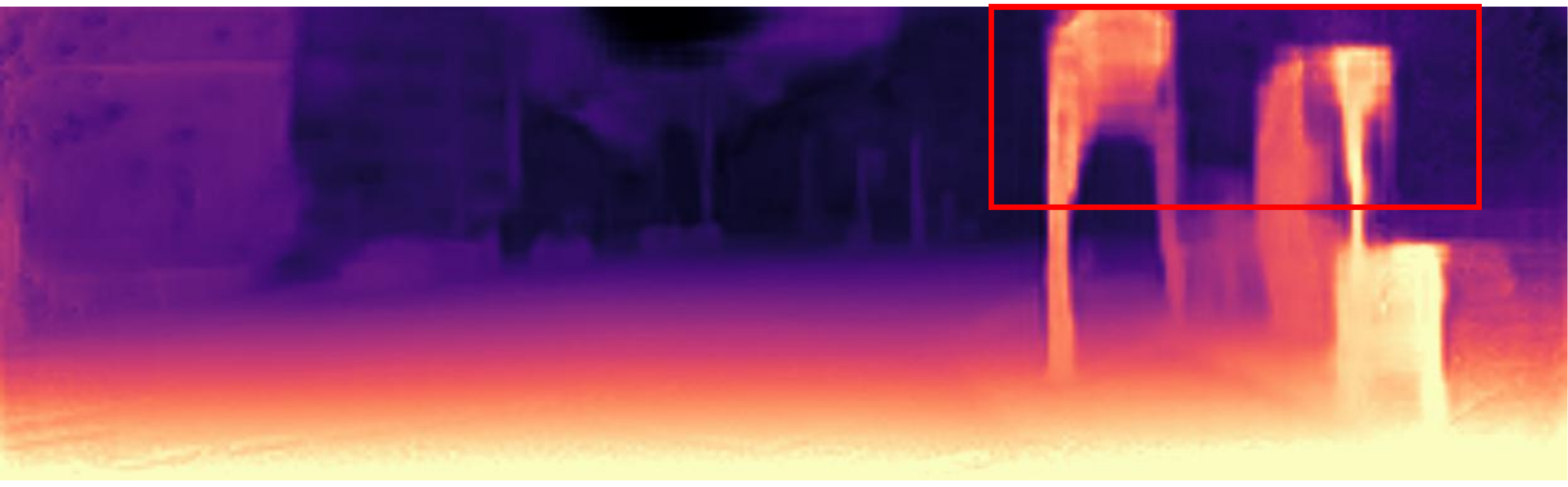}}
\end{center}
   \caption{An example of mosaic artifact when using CWS in the last fusion block.}
\label{Fig:ChoFus}
\end{figure}

\noindent\textbf{Different choices of fusion block.} We introduced two different fusion blocks in Sec.~\ref{Sec:FusBlock}. In this section, we test each type of fusion block and show the effects it brings. We set the $sc$ and $pc$ to $32$ and $32$, respectively. For the last fusion block (oFus) and other fusion blocks (Fus), we test different combinations of these two fusion strategies. For CWS fusion, we also report the results of the weighted and the un-weighted version. Table~\ref{Tab:ChoFus} shows the results in which we can see that the combination of CWS and CTC gets the best results. In this experiment, we found using CWS in the last fusion block will bring the mosaic artifact to the prediction. As shown in Figure~\ref{Fig:ChoFus}, this artifact appears in the edges of objects like poles and traffic signs. We speculate this is caused by the upsampling and addition operation. By using the CTC in the last fusion block, we can effectively eliminate this artifact.

\begin{table}[htbp]
  \centering
  \footnotesize
  \setlength{\tabcolsep}{4.5pt}
  \caption{Results of different combinations of fusion blocks. WS denotes the weighted CWS and S denotes the un-weighted one. C denotes the CTC.}
  \vspace{2pt}
    \begin{tabular}{cc|ccccc}
    \hline
    Fus & oFus &\cellcolor[rgb]{.988,.894,.839} Abs REL &\cellcolor[rgb]{.988,.894,.839} Sq REL &\cellcolor[rgb]{.988,.894,.839} RMSE  &\cellcolor[rgb]{.988,.894,.839} $\mathrm{RMSE}_{\log}$ &\cellcolor[rgb]{.886,.937,.855} $\delta_1$ \\
    \hline
    \hline
    WS & WS & 0.115 & 0.800 & 4.597 & 0.190 & 0.872 \\
    C  & C  & 0.114 & 0.794 & 4.583 & 0.189 & 0.872 \\
    WS & C  & 0.112  & 0.776  &  4.574 &  0.188  & 0.875 \\
    S  & C  & 0.115 & 0.811 & 4.700 & 0.190 & 0.874 \\
    \hline
    \end{tabular}%
  \label{Tab:ChoFus}%
\end{table}%

\begin{table*}[htbp]
  \centering
  \footnotesize
  \setlength{\tabcolsep}{2.0pt}
  \caption{Quantitative results. Comparing with FCN and Deeplabv3, our model gets the best performance around many classes.}
    \begin{tabular}{l|c|ccccccccccccccccccc|c}
    \hline
    \rowcolor[rgb]{.867,.922,.969}\cellcolor[rgb]{1.,1.,1.} Network & \cellcolor[rgb]{1.,1.,1.}Params& road  & sdwk  & bldng & wall  & fence & pole  & light & sign  & veg   & trrn  & sky   & psn   & rider & car   & truck & bus   & train & moto  & bike  & \cellcolor[rgb]{.886,.937,.855}mIoU \\
    \hline
    \hline
    FCN (ResNet50) & 33.0M & 89.0 & 31.6 & 78.9 &22.5& 19.7 & 20.1 & 13.8 & 6.72 & 77.1 & 21.4 &83.3& 40.7 &\textbf{9.79}& 76.0 & 24.8 & 13.2  &\textbf{8.48}& 12.9 & 0.00 & 34.21 \\
    Deeplabv3 (ResNet50) & 39.6M & 85.4  & 30.2  & 78.8  &\textbf{23.0}&  13.3 & 20.1  & 13.5  & 3.86  & 77.3 & 19.9 &\textbf{83.4}& 39.4 & 4.93 & 76.4 &\textbf{26.9}& \textbf{16.8} & 0.98 &\textbf{14.2}& 0.00 & 33.07 \\
    \hline
    \rowcolor[rgb]{.851,.851,.851}PFN (sc=48, pc=144) & 34.1M &\textbf{90.1}&\textbf{39.2}&\textbf{82.0}& 22.4 &\textbf{20.1}&\textbf{37.9}&\textbf{26.2}&\textbf{9.07}&\textbf{81.8}&\textbf{27.4}& 80.8 & \textbf{46.6}& 6.13 & \textbf{81.2} & 22.8 & 11.5 & 0.00 & 11.4 & 0.00 &\textbf{36.67}\\
    \hline
    \end{tabular}%
  \label{Tab:SemSeg}%
\end{table*}%

\begin{figure*}
\begin{center}
\scalebox{0.3}[0.3]{\includegraphics{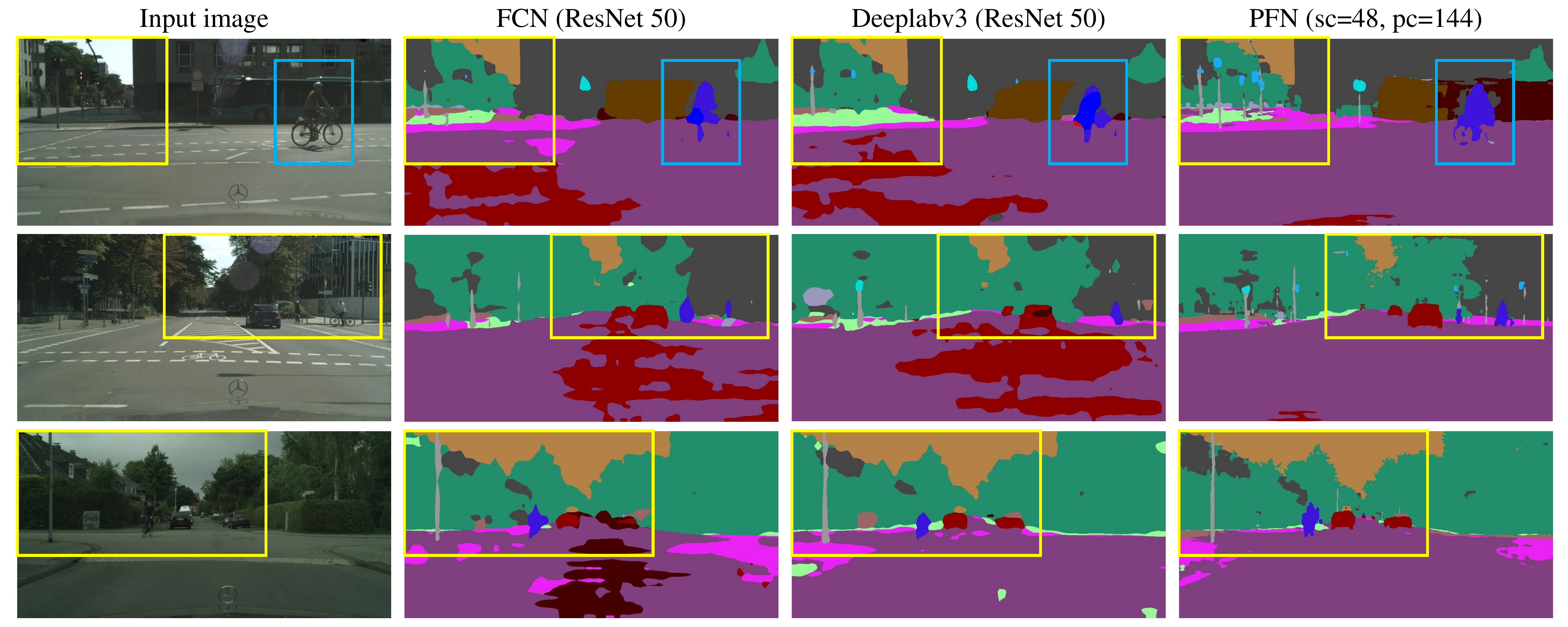}}
\end{center}
   \caption{Qualitative results. Our model predicts more refined boundaries and learns more global structure information.}
\label{Fig:SemSeg}
\end{figure*}

\section{Experiments of Domain Randomization on Semantic Segmentation}

To demonstrate the ability of handle other supervised pixel-wise prediction tasks and the ability of generalization across dataset domains, we test PFNs on the task of domain randomization on semantic segmentation. On this task, the segmentation network is trained without accessing any data of the target domain and tested on the unseen target domain. We select the virtual dataset GTA5~\cite{richter2016playing} for training and the real dataset Cityscapes~\cite{cordts2016cityscapes} for testing in these experiments.

\subsection{Implementation Details}
The proposed PFN is utilized straightforwardly as the segmentation network. We modify the output channel of the last convolution layer to the number of classes of the target dataset. The hyper-parameters of $S, sc, pc$ are set to $5, 48, 144$ respectively. We calculate losses from outputs of scales and average the losses for training. Following~\cite{chen2017rethinking}, we set the learning rate to $1\times 10^{-2}$ and employ a poly learning rate policy where the initial learning rate is multiplied by $(1-\frac{\mathrm{iter}}{\mathrm{max\_iter}})^{power}$ where $power=0.9$. We resize the input images to $512\times 256$. When testing, the output segmentation map is upsampled to calculate the mIoU metric. We set the max iteration to $250000$ and set the batch size to $2$. For data augmentation, we randomly left-right flipping the input images and apply the \verb'ColorJitter' and \verb'RandomBlur' operations to expand the training data distribution.
\subsection{Results}

We compare our methods with FCN~\cite{long2015fully} and Deeplabv3~\cite{chen2017rethinking}. We employ ResNet50 for their backbone and use the same training settings to report the results. As shown in Table~\ref{Tab:SemSeg}, our PFN gets the best mIoU compared with the other two methods. Zoom in to each class, the PFN gets better predictions on the background classes like sidewalk and building. The thin pole is also predicted well. Figure~\ref{Fig:SemSeg} illustrates the qualitative comparison between those methods. We can see that our PFN keeps the global structure and predicts the best edges between the sky and the plants. Yue~\etal~\cite{yue2019domain} introduce another method for domain randomization on semantic segmentation. Our method would be complementary to theirs.

\section{Conclusion}

In this paper, we propose the fractal pyramid networks as an alternative to the encoder-decoder structure. In the PFNs, we show another way for information encoding. Rather than try to encode signals to a large-channel feature, we encode information to many separate small-channel features by multiple information processing pathways. We experimentally validate our networks on two pixel-wise prediction tasks and show that our networks get competitive results comparing with the encoder-decoder networks with better detail recovery and better temporal consistency, which clearly demonstrate the effectiveness of our networks.

{\small
\bibliographystyle{ieee_fullname}
\bibliography{egbib}
}

\end{document}